\documentclass[journal]{IEEEtran}  

\usepackage{amsmath,graphicx}
\usepackage{algorithm}
\usepackage[]{graphicx}
\usepackage{stfloats}
\usepackage{cite}
\usepackage{psfrag}
\usepackage{wrapfig}
\usepackage{epsfig}
\usepackage[hyphens]{url}
\usepackage{amsmath, amssymb, bm}
\usepackage{array}
\usepackage{times}
\usepackage{multicol}
\usepackage{algorithm}
\usepackage{algorithmic}
\usepackage{mathrsfs}
\usepackage{multirow}
\usepackage{txfonts}
\usepackage[normalem]{ulem}

\usepackage[utf8x]{inputenc}
\usepackage{ucs}
\usepackage{makeidx}
\usepackage[dvipsnames,usenames]{color}
\usepackage{array}
\usepackage{colortbl}
\usepackage{booktabs}
\usepackage{color}
\usepackage{cleveref}
\usepackage{setspace}
\usepackage{enumitem} 
\usepackage{soul} 
\usepackage{pifont} 
\usepackage{balance}

\DeclareMathOperator*{\argmax}{arg\,max}
\renewcommand{\vec}[1]{\mathbf{#1}}
\newcommand{\setsym}[1]{\mathbb{#1}}
\newcommand*{\vertbar}{\rule[-0.5ex]{0.5pt}{2.5ex}} 

\graphicspath{ {./fig/} }
\renewcommand{\baselinestretch}{1.0}

\begin{document}
\title{Advancing Acoustic-to-Word CTC Model with Attention and Mixed-Units}

\author{Amit~Das,~\IEEEmembership{Student~Member,~IEEE,}
        Jinyu~Li,~\IEEEmembership{Member,~IEEE,}
        Guoli Ye,~\IEEEmembership{Member,~IEEE,}
        Rui~Zhao,~\IEEEmembership{Member,~IEEE,}
        and~Yifan~Gong,~\IEEEmembership{Senior~Member,~IEEE}
\thanks{The authors are with Microsoft Corporation, USA (email: amitdas@illinois.edu; jinyli@microsoft.com; guoye@microsoft.com; ruzhao@microsoft.com; yifan.gong@microsoft.com).}
}

%


\maketitle
\IEEEpeerreviewmaketitle 

\begin{abstract}
The acoustic-to-word model based on the Connectionist Temporal Classification (CTC) criterion is a natural end-to-end (E2E) system directly targeting word as output unit. Two issues exist in the system: first, the current output of the CTC model relies on the current input and does not account for context weighted inputs. This is the hard alignment issue. Second, the word-based CTC model suffers from the out-of-vocabulary (OOV) issue. This means it can model only frequently occurring words while tagging the remaining words as OOV. Hence, such a model is limited in its capacity in recognizing only a fixed set of frequent words. In this study, we propose addressing these problems using a combination of attention mechanism and mixed-units. In particular, we introduce Attention CTC, Self-Attention CTC, Hybrid CTC, and Mixed-unit CTC. 

First, we blend attention modeling capabilities directly into the CTC network using Attention CTC and Self-Attention CTC. Second, to alleviate the OOV issue, we present Hybrid CTC which uses a word and letter CTC with shared hidden layers. The Hybrid CTC consults the letter CTC when the word CTC emits an OOV. Then, we propose a much better solution by training a Mixed-unit CTC which decomposes all the OOV words into sequences of frequent words and multi-letter units. Evaluated on a 3400 hours Microsoft Cortana voice assistant task, our final acoustic-to-word solution using attention and mixed-units achieves a relative reduction in word error rate (WER) over the vanilla word CTC by 12.09\%. Such an E2E model without using any language model (LM) or complex decoder also outperforms a traditional context-dependent (CD) phoneme CTC with strong LM and decoder by 6.79\% relative.
\end{abstract}


%
\IEEEpeerreviewmaketitle

\section{Introduction}
\label{sec: Intro}
\IEEEPARstart{I}{n} automatic speech recognition (ASR), we are given a sequence of acoustic feature vectors $\vec{x}$. The objective is to decode a sequence of words $\vec{y}$ from $\vec{x}$ with minimum probability of error. With the 0-1 loss function, the optimal solution uses the Bayesian Maximum Aposteriori (MAP) rule
\begin{align}
\hat{\vec{y}} &=  \argmax_{\vec{y}} \ P(\vec{y}|\vec{x}; \Theta_{\text{ASR}}), \label{eq:asr_map} \\ 
              &=  \argmax_{\vec{y}} \ P(\vec{x}|\vec{y}; \Theta_{\text{AM}}) P(\vec{y}; \Theta_{\text{LM}}). \label{eq:asr_am_lm}
\end{align}
However, to reduce complexity, practical ASR systems often use the sub-optimal solution
\begin{align}
\hat{\vec{y}} &\approx \argmax_{\vec{y, l}} \  P(\vec{x}|\vec{l}; \Theta_{\text{AM}}) P(\vec{l}|\vec{y}; \Theta_{\text{PM}}) P(\vec{y}; \Theta_{\text{LM}}). \label{eq:asr_am_pm_lm}
\end{align}
Here, $\vec{l}$ is a sequence of phonemes and $\Theta_{\text{ASR}} = \{\Theta_{\text{AM}}, \Theta_{\text{PM}}, \Theta_{\text{LM}}\}$ is the set of parameters to be estimated during training. The first term $P(\vec{x}|\vec{l}; \Theta_{\text{AM}})$ in Eq.~\eqref{eq:asr_am_pm_lm} is the likelihood of the features given the phoneme sequence and is obtained from an acoustic model (AM). The second term $P(\vec{l}|\vec{y};\Theta_{\text{PM}})$ is the likelihood of the phoneme sequence given the word sequence and is obtained from a lexicon or pronunciation model (PM). The third term $P(\vec{y}; \Theta_{\text{LM}})$ is the prior probability of the word sequence and is obtained from a language model (LM). 

In theory, all $\{\Theta_{\text{AM}}, \Theta_{\text{PM}}, \Theta_{\text{LM}}\}$ should be estimated jointly. However, in practice, they are estimated separately and hence training an ASR system becomes a complex disjoint learning problem. Moreover, decoding at test time involves a complex graph search procedure which is intensive both in time and memory. This makes traditional ASR systems often cumbersome for deployment in real-world devices.

In contrast, an end-to-end (E2E) ASR system \cite{Yu-RecentProgDeepLearningAcousticModels, sak2015learning, miao2015eesen, Chan-LAS, prabhavalkar2017comparison, battenberg2017exploring, sak2017recurrent, hadiantowards, chiu2018state, sainath2017improving} circumvents the disjoint learning problem by directly transducing a sequence of features $\vec{x}$ to a sequence of words $\vec{y}$. Some widely used contemporary neural network based E2E approaches for sequence-to-sequence transduction are: (a) Connectionist Temporal Classification (CTC) \cite{Graves-CTCFirst, Graves-E2EASR}, (b) Recurrent Neural Network (RNN) Encoder-Decoder (ED) \cite{Cho-RNNEncDecSMT, Bahdanau-RNNEncDecAlignTranslate, Bahdanau-AttentionASR, Chorowski-AttentionASR}, and (c) RNN Transducer (RNN-T) \cite{Graves-RNNSeqTransduction}. These approaches have been successfully applied to large scale ASR \cite{sak2015learning, miao2015eesen, Lu2015StudyRNNED, Chan-LAS, soltau2016neural, prabhavalkar2017comparison, battenberg2017exploring, rao2017exploring, chiu2018state, masumura2019largecontext,moritz2019triggeredattention,bahar2019onusing2ds2s,xiang2019crfbased}. In this study, we confine ourselves to the CTC approach.

CTC, first introduced in \cite{Graves-CTCFirst, Graves-E2EASR}, involves training a stack of underlying RNNs and minimizing the sequence level cross-entropy (CE) loss $-\text{log}\ P(\vec{y}|\vec{x})$. In contrast, RNN training minimizes the frame level CE loss. Moreover, CTC networks offer the versatility to model output units of different sizes such as monophones, characters, words, or other sub-word units. Owing to this simplicity in the training structure and versatility of output units, CTC is regarded as one of the most popular E2E methods \cite{Hannun-DeepSpeech, sak2015learning, sak2015fast, miao2015eesen, kanda2016maximum, soltau2016neural, Zweig-AdvancesNeuralASR, liu2017gram, audhkhasi2017direct, Li17CTCnoOOV, Yu-RecentProgDeepLearningAcousticModels, Li2018Speaker}. 

In ASR, the number of output labels in $\vec{y}$ is usually smaller than the number of input speech frames in $\vec{x}$. However, since a CTC network is essentially an RNN, it is forced to predict a label for every frame in $\vec{x}$. Since some frames may not always be associated with a label (a) CTC introduces a special \textit{blank} label as an additional output label which acts as a filler and, (b) it allows for repetition of labels (for both blank or non-blank). As a result, CTC frame level outputs are usually dominated by blank labels. The outputs corresponding to the non-blank labels usually occur with spikes in their posteriors because of their high confidences. Thus, an easy way to convert intermediate frame level outputs to final ASR outputs using CTC involves a simple two-step procedure. In the first step, generate a sequence of labels corresponding to the highest posteriors and merge consecutive duplicate labels. In the second step, remove the blank labels and concatenate the remaining non-blank labels into words. This is known as greedy decoding. It is a very attractive feature for E2E modeling as there is neither any LM nor any complex decoding involved. This makes it easy for deployment in real-world devices. The E2E ASR developed in this study uses greedy decoding.

As the goal of ASR is to generate a word sequence from speech acoustics, the word is the most natural output unit compared to other output units such as monophones or characters. A big challenge in the word-based CTC model, a.k.a. acoustic-to-word (A2W) CTC or word CTC, is the OOV issue \cite{bazzi2002modelling, decadt2002transcription, yazgan2004hybrid, bisani2005open}. In \cite{sak2015fast, soltau2016neural, audhkhasi2017direct}, only the most frequent words in the training set were used as output targets whereas the remaining words were lumped together as OOVs. These OOVs can neither be modeled nor recognized correctly. For example, consider an utterance containing the sequence ``have you been to newyorkabc" in which ``newyorkabc" is an OOV (infrequent) word. For an OOV-based model, a likely output for this utterance would be ``have you been to OOV''. Despite it being the expected output from the OOV-based model, the presence of the OOV tag in the sentence degrades the end-user experience. Another disadvantage of OOV modeling is that the data related to those infrequent words are wasted, resulting in reduced modeling power. To underscore this issue, \cite{sak2015fast} trained a word CTC with up to 25 thousand (k) word targets. However, the ASR accuracy of the word CTC was far below the accuracy of a context dependent (CD) phoneme CTC model with LM, partially due to the high OOV rate when using only around 3k hours of training data. 

The accuracy gap between a word CTC and CD phoneme CTC can be attributed to multiple reasons. First, training a word CTC requires orders of magnitude of more training data than a CD phoneme CTC because words which qualify as non-OOVs (frequent words) require sufficient number of training examples. Words which do not meet this sufficiency requirement are simply tagged as OOVs. Hence, such words can neither be modeled as valid words during training nor recognized during evaluation. Second, even in the presence of large training data, it is difficult to capture the entire vocabulary of a language. For example, a word CTC cannot handle unfamiliar nouns or emerging hot-words (e.g. selfie, meme, unfriend) which gradually become popular after an acoustic model has been built.

Several studies in the past have attempted to address these issues. In \cite{soltau2016neural}, it was shown that by using 100k words as output targets and by training the model with 125k hours of data, a word CTC was able to outperform a CD phoneme CTC. However, easy accessibility to such large databases is rare. Usually, at most a few thousand hours of data are available. In \cite{audhkhasi2018building}, the authors were able to train a word CTC model with only 2k hours of data achieving ASR accuracy comparable to that of a CD phoneme CTC. Their proposed training regime included initializing the word CTC with a well-trained phoneme CTC, curriculum learning \cite{Bengio2009Curriculum}, Nesterov momentum-based stochastic gradient descent, dropout, and low rank matrix factorization \cite{Sainath2013LRMF}. To address the hot-words issue, \cite{audhkhasi2018building} also proposed a spell and recognize (SAR) model which has a combination of words and characters as output targets. The SAR model is used to learn to first spell a word as a sequence of characters and then recognize it as a whole word. However, whenever an OOV is detected, the decoder consults the letter sequence from the speller. Thus, the displayed hypothesis to the end-user contains words (for non-OOVs) and characters (for OOVs). Spelling out the characters for OOVs is more meaningful to the users than simply displaying ``OOV". However, it was reported that the overall recognition accuracy of the SAR model improved only marginally over a word-only CTC. In \cite{Chen2018OnModular}, the authors proposed training two CTC models separately - an acoustics-to-phoneme model from acoustic data and a phoneme-to-word model using text data respectively. Then, the two models were jointly optimized resulting in an A2W model. 

In this study, we propose four solutions to improve the recognition accuracy of the all-neural word CTC using only 3400 hours of training data while also alleviating the OOV issue.
\begin{itemize}[leftmargin=*]
\item First, in Section \ref{sec: CTCAttn}, we propose \textit{Attention CTC} \cite{Das18CTCAttention} to address the inherent hard alignment problem in CTC. Since CTC relies on the hidden feature vector at the current time to make predictions, it does not directly attend to feature vectors of the neighboring frames. This is the hard alignment problem which makes CTC's output independent assumption worse. Our proposed solution generates new hidden features that carry attention weighted context information. We achieved this by blending some concepts from RNN-ED into CTC modeling.

\item Second, in Section~\ref{sec: SelfAttnCTC}, we investigate another attention mechanism called \textit{Self-Attention} \cite{vaswani2017selfattention} in CTC networks.

\item Third, we propose \textit{Hybrid CTC} \cite{Li17CTCnoOOV} which is a single CTC consisting of a word CTC and a letter CTC trained jointly using multi-task learning (MTL) \cite{Caruana-MTL,Seltzer-MTLPhonemeRecog}. We train the word CTC first and then add a letter CTC as an auxiliary task by sharing the hidden layers of the word CTC. During recognition, the word and letter CTCs generate sequences of words and letters respectively. However, the letter CTC is consulted for the letter sequence only when the word CTC emits an OOV token. This makes the Hybrid CTC capable of recognizing OOVs and thereby reducing errors introduced by OOVs.

\item Finally, we further improve the word CTC and reduce OOV errors by introducing \textit{Mixed-unit CTC} \cite{Li18CTCnoOOV}. Here, during training, the OOV word is decomposed into a sequence of frequent words and letters (which we refer to as \textit{mixed-units}). During testing, we perform greedy decoding for the whole E2E system in a single step without the need of using the two-stage process (OOV-detection and then letter-sequence-consulting) as in Hybrid CTC. We will later show that a CTC with mixed-units outperformed a CTC with wordpieces which have become popular in recent RNN-ED frameworks \cite{chiu2018state}. 
\end{itemize}

Our final proposed word CTC achieved a relative WER reduction (WERR) of about 12.09\% over the vanilla word CTC \cite{Graves-CTCFirst}. Furthermore, the same word CTC outperformed the traditional CD phoneme CTC with a strong LM and decoder by 6.79\% relative.

The remainder of the article is organized as follows. In Section~\ref{sec: E2E} we give a brief overview of CTC and RNN-ED. In Sections~\ref{sec: CTCAttn}, \ref{sec: SelfAttnCTC}, \ref{sec: hybCTC}, \ref{sec: multimixCTC}, we explain the proposed  Attention CTC, Self-Attention CTC, Hybrid CTC, and Mixed-unit CTC respectively. In Section~\ref{sec: Expts}, we provide experimental evaluations of our proposed algorithms. Finally, we summarize our study and draw conclusions in Section~\ref{sec: Conclusions}. The terms letter and character have been interchangeably used in this study.

\section{End-to-End Speech Recognition}
\label{sec: E2E}
An E2E ASR system models the posterior distribution $p(\vec{y}|\vec{x})$ by transducing an input sequence of acoustic feature vectors $\vec{x}$ to an output sequence of tokens $\vec{y}$ (phonemes, characters, words etc.). More specifically, for an input sequence of feature vectors $\vec{x} = (\vec{x}_{1}, \cdots, \vec{x}_{T})$ of length $T$ with $\vec{x}_{t} \in \setsym{R}^{m}$, an E2E ASR system transduces the input sequence to an intermediate sequence of hidden feature vectors $\vec{h} = (\vec{h}_{1}, \cdots, \vec{h}_{L})$ of length $L$ with $\vec{h}_{l} \in \setsym{R}^{n}$. The sequence $\vec{h}$ undergoes another transduction resulting in an output sequence $\vec{y}$ whose posterior probability is $\tilde{p}(\vec{y}|\vec{x})$. Here $\vec{y} = (y_{1}, \cdots, y_{U})$ is of length $U$ with $y_{u} \in \setsym{L}$, $\setsym{L}$ being the label set. Usually $U \leq T$ and $L = T$ in E2E ASR systems. Thus, an E2E neural network, parameterized by $\mathbf{W}$, learns a many-to-one functional $\vec{f}_{\mathbf{W}}: \vec{x} \mapsto \tilde{p}(\vec{y}|\vec{x})$ where $\tilde{p}(\vec{y}|\vec{x})$ closely resembles the true $p(\vec{y}|\vec{x})$.

\vspace{-2mm}
\subsection{Connectionist Temporal Classification (CTC)}
\label{ssec: CTC}
A CTC network uses a recurrent neural network (RNN) and the CTC error criterion \cite{Graves-CTCFirst, Graves-E2EASR}  which directly optimizes the prediction of a transcription sequence. As the length of the output labels is shorter than the length of the input speech frames, a CTC path is introduced to make their lengths equal by adding the blank symbol $\phi$ as an additional label and allowing repetition of labels. Thus, the new label set becomes $\setsym{L}^{\prime} = \setsym{L} \cup \phi$. Let $K = \left\vert\setsym{L}^{\prime}\right\vert$ be the cardinality of the label set $\setsym{L}^{\prime}$.

Denote $\bm\pi = (\pi_{1}, \cdots, \pi_{T})$ as the CTC path (or alignment) with $\pi_{t} \in \setsym{L}^{\prime}$, $\bf{y}$ as the target label sequence (transcription) we want to recognize, and  $B^{-1}(\bf{y})$ as the preimage of $\vec{y}$ mapping all possible CTC paths $\bm\pi$ that result in $\bf{y}$. Then, the CTC loss function is defined as the negative log of sum of the probabilities of all possible CTC paths $\bm\pi$ that result in $\bf{y}$. This is given by
\begin{equation}
L_{CTC} = - \ln p( {\bf{y}|\bf{x}} ) = - \ln \sum_{{\bm\pi} \in B^{-1}(\bf{y})} p( {\bm\pi}  | \bf{x} ) \label{eq:ctcloss}.
\end{equation}
With the conditional independence assumption ($\pi_{t} \Perp \pi_{\ne t}|\vec{x}$), $p( {\bm\pi} | \bf{x} )$ can be further decomposed into a product of posteriors of each frame as \vspace{-1mm}
\begin{equation}
p( {{\bm\pi}  | \bf{x}} ) = \prod_{t=1}^T p( \pi_{t}| \bf{x}) \label{eq:ctcpathprob}.
\end{equation}
During decoding, it is very simple to generate the decoded sequence using greedy decoding: simply concatenate the labels corresponding to the highest posteriors and merge the duplicate labels; then remove the blank labels. Thus, there is neither a language model nor any complex graph search in greedy decoding.

\subsection{RNN Encoder-Decoder (RNN-ED)}
\label{ssec: RNNEncDec}
An RNN-ED \cite{Cho-RNNEncDecSMT, Bahdanau-RNNEncDecAlignTranslate, Bahdanau-AttentionASR, Chorowski-AttentionASR} uses two distinct networks - an RNN encoder network that transforms $\vec{x}$ into $\vec{h}$ and an RNN decoder network that transforms $\vec{h}$ into $\vec{y}$. Using these, an RNN-ED models $p(\vec{y}|\vec{x})$ as
\vspace{-2mm}
\begin{align}
p(\vec{y}|\vec{x}) &= \prod_{u=1}^{U} p(y_{u}|\vec{y}_{1:u-1}, \vec{c}_{u}), \label{eq:RNNED-transcriptprob}
\end{align}
where $\vec{c}_{u}$ is the context vector at time $u$ and is a function of $\vec{x}$.
There are two key differences between CTC and RNN-ED. First, $p(\vec{y}|\vec{x})$ in Eq.~\eqref{eq:RNNED-transcriptprob} is generated using a product of ordered conditionals. Thus, RNN-ED is not impeded by the conditional independence constraint of Eq.~\eqref{eq:ctcpathprob}. 
Second, The decoder output $y_{u}$ at time $u$ is dependent on $\vec{c}_{u}$ which is a weighted sum of all its inputs (soft alignment), i.e., $\vec{h}_{t}, t = 1, \cdots, T$. In contrast, CTC generates $y_{u}$ using only $\vec{h}_{t}$ (hard alignment).

The decoder network of RNN-ED has three components: a multinomial distribution generator Eq.~\eqref{eq:RNNED-generate}, an RNN decoder Eq.~\eqref{eq:RNNED-recurrent}, and an attention network Eq.~\eqref{eq:RNNED-annotate}-\eqref{eq:RNNED-locfeat} \cite{Chorowski-AttentionASR, Bahdanau-AttentionASR} as follows:
\begin{align}
p(y_{u}|\vec{y}_{1:u-1}, \vec{c}_{u}) &= \text{Generate}(y_{u-1}, \vec{s}_{u}, \vec{c}_{u}), \label{eq:RNNED-generate} \\
\vec{s}_{u} &= \text{Recurrent}(\vec{s}_{u-1}, \vec{y}_{u-1}, \vec{c}_{u}), \label{eq:RNNED-recurrent} \\
\vec{c}_{u} &= \text{Annotate}(\bm\alpha_{u}, \vec{h}) = \sum_{t=1}^{T} \alpha_{u,t} \vec{h}_{t}, \label{eq:RNNED-annotate} \\
\alpha_{u,t} &= \text{Attend}(\vec{s}_{u-1}, \bm\alpha_{u-1}, \vec{h}_{t}), \quad t = 1, \cdots, T. \label{eq:RNNED-attend}
\end{align}
Here, $\vec{h}_{t}, \vec{c}_{u} \in \setsym{R}^{n}$, and $\bm\alpha_{u} = [\alpha_{u,1} \cdots \alpha_{u,T}]$ is a probability distribution. Hence, $\alpha_{u,t} \in \setsym{U}$ with $\setsym{U} = [0,1]$ such that $\sum_t \alpha_{u,t} = 1$. Also, for simplicity assume $\vec{s}_{u} \in \setsym{R}^{n}$. $\text{Generate}(.)$ is a feedforward network with a softmax operation generating the ordered conditional $p(y_{u}|\vec{y}_{1:u-1}, \vec{c}_{u})$ \vphantom{for my reference only,see \cite[Appendix A.2.2]{Bahdanau-RNNEncDecAlignTranslate}}.  Recurrent(.) is an RNN decoder operating on the output time axis indexed by $u$ and has hidden state $\vec{s}_{u}$. Annotate(.) computes the context vector $\vec{c}_{u}$ (also called the soft alignment) using the attention probability vector $\bm\alpha_{u}$ and the hidden sequence $\vec{h}$. Attend(.) computes the attention weight $\alpha_{u,t}$ using a single layer feedforward network (Score(.) function) followed by softmax normalization as follows:
\begin{align}
e_{u,t} &= \text{Score}(\vec{s}_{u-1}, \bm\alpha_{u-1}, \vec{h}_{t}), \quad t = 1, \cdots, T, \label{eq:RNNED-score} \\
\alpha_{u, t} &= \frac{ \text{exp}(e_{u, t}) } { \sum_{t^{\prime}=1}^{T} \text{exp}(e_{u, t^{\prime}}) }, \quad t = 1, \cdots, T. \label{eq:RNNED-normalizedscore}
\end{align}
Here, $e_{u, t} \in \setsym{R}$ and $\text{Score}(.)$ can either be a content-based or hybrid-based function. The latter encodes both content ($\vec{s}_{u-1}$) and location ($\bm\alpha_{u-1}$) information. $\text{Score}(.)$ is computed using
\begin{align}
\hspace{-2mm} e_{u, t} &= \begin{cases}
\vec{v}^{T}\text{tanh}\ (\vec{U} \vec{s}_{u-1} + \vec{W} \vec{h}_{t}  +  \vec{b}), \ \mbox{(content)} \\
\vec{v}^{T}\text{tanh}\ (\vec{U} \vec{s}_{u-1} + \vec{W} \vec{h}_{t}  + \vec{V} \vec{f}_{u} + \vec{b}), \ \mbox{(hybrid)}
\end{cases} \label{eq:RNNED-ContentHybrid} \\
&\text{where,} \quad \vec{f}_{u} = \vec{F} \ast \bm\alpha_{u-1} \label{eq:RNNED-locfeat}.
\end{align}
The operation $\ast$ denotes convolution. Thus, in the hybrid case, the dependence on $\bm\alpha_{u-1}$ is through $\vec{f}_{u}$. Attention parameters $\vec{U}, \vec{W}, \vec{V}$, $\vec{F}, \vec{b}, \vec{v}$ are learned while training RNN-ED.

\section{Attention CTC}
\label{sec: CTCAttn}

\begin{figure*}[ht]
\centering
\resizebox{0.90\linewidth}{!}{
\includegraphics[width=\textwidth,height=0.48\textwidth,trim=4mm 0mm 2mm 2mm,clip]{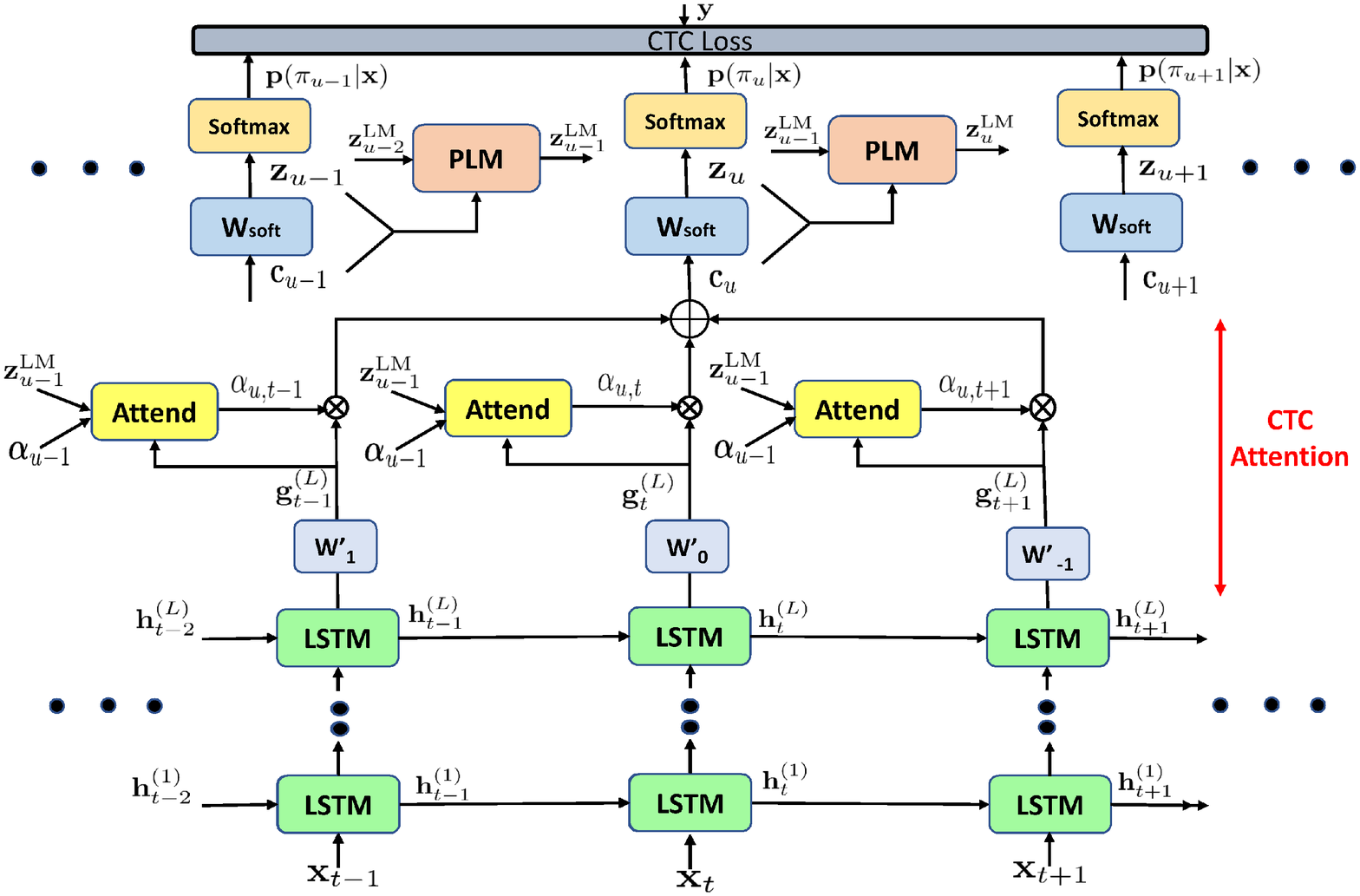}\vspace{-2mm}
}
\caption{An example of an Attention CTC network with an attention window of size $C = 3$.}
\label{fig:CTCAttn}\vspace{-4mm}
\end{figure*}

In this section, we outline various steps required to model attention directly within CTC. In the past, several attempts have been made to apply attention on E2E models. For example, attention-based RNN-ED \cite{Chorowski-AttentionASR, Bahdanau-AttentionASR} network was used to predict word outputs in \cite{lu2016training}. Other studies have investigated using CTC as an auxiliary task to improve attention-based RNN-ED using an MTL framework. For example, CTC was used either at the top layer \cite{Kim-JointCTCRNNEncDecUsingMTL, hori2017advances} or at an intermediate layer \cite{Toshniwal-MTLLowLevelRNNED} in the MTL framework. Extensions of CTC such as RNN-T \cite{Graves-RNNSeqTransduction, rao2017exploring} and RNN aligner \cite{sak2017recurrent} either change the objective function or the training process to relax the frame independence assumption of CTC. However, none of these approaches used attention directly within the CTC network. The proposed Attention CTC model is different from all these approaches since we use attention mechanism to improve the hidden layer representations with more context information without changing the CTC objective function and the training process. Our primary motivation in this work is to address the hard alignment problem of CTC, as outlined earlier in Section~\ref{sec: Intro}, by modeling attention directly within the CTC framework.

An example of the proposed Attention CTC network is shown in Figure \ref{fig:CTCAttn}.  We propose the following key ideas to blend attention into CTC. (a) First, we derive context vectors using \textit{time convolution features} (Section \ref{ssec: CTCAttn-conv}) and apply attention weights on these context vectors (Section \ref{ssec: CTCAttn-attn}). This makes it possible for CTC to be trained using soft alignments instead of hard. (b) Second, to improve attention modeling, we incorporate a \textit{pseudo language model} (Section \ref{ssec: CTCAttn-LM}) during CTC training. (c) Finally, we improve our attention modeling further by introducing \textit{component attention} (Section \ref{ssec: CTCAttn-Comp}) where context vectors are produced as a result of applying attention on hidden features across both time and their individual components.
We explain each of these ideas separately with illustrations in the following subsections.
We will use the indices $t$ and $u$ to denote the time step for input $\vec{h}$ and output $\vec{c}$ respectively of the attention block to maintain notational consistency with RNN-ED. 

\vspace{-3mm}
\subsection{Time Convolution (TC) Features}
\label{ssec: CTCAttn-conv}
\begin{figure}[th]
\centering
\resizebox{0.90\linewidth}{!}{
\includegraphics[width=\textwidth,height=0.48\textwidth,trim=4mm 0mm 2mm 2mm,clip]{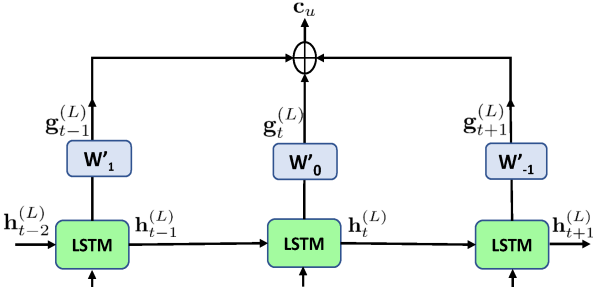}\vspace{-2mm}
}
\caption{Time convolution with an attention window of size $C = 3 \ (\text{i.e., } \tau = 1)$.}
\label{fig:TC}\vspace{-3mm}
\end{figure}

First, we construct TC features from the hidden outputs $\vec{h}$ of the last LSTM layer. This is illustrated in Fig.~\ref{fig:TC}.  Consider a subsequence of $\vec{h}$ rather than the entire sequence. We refer to this subsequence, $(\vec{h}_{u-\tau}, \cdots, \vec{h}_{u}, \cdots, \vec{h}_{u+\tau})$, as the \textit{attention window}. Each $\vec{h}_{t} \in \setsym{R}^{n}$.  The attention window is centered around the current time $u$ with $\tau$ being the length of the attention window on either side of $u$. Thus, the total length of the attention window is $C = 2\tau + 1$. Now consider $C$ time convolution kernels $(\vec{W}^{\prime}_{u-\tau}, \cdots, \vec{W}^{\prime}_{u}, \cdots, \vec{W}^{\prime}_{u+\tau})$ where $\vec{W}^{\prime}_{t} \in \setsym{R}^{n \times n}$ and $\vec{W}^{\prime}_{t_1} \ne \vec{W}^{\prime}_{t_2}$ for $t_{1} \ne t_{2}$. Then the context vector $\vec{c}_{u}$ is computed using time convolution as,
\begin{align}
\vec{c}_{u} 
            & = \sum_{t =u-\tau}^{u+\tau} \vec{W}^{\prime}_{u - t} \vec{h}_{t} \nonumber \\
			&\stackrel{\Delta}{=} \sum_{t =u-\tau}^{u+\tau} \vec{g}_{t} \nonumber \\
			&= \gamma \sum_{t =u-\tau}^{u+\tau} \alpha_{u,t} \vec{g}_{t}. \label{eq:CTCAttn-TimeConvolution}
\end{align}
Here, $\vec{g}_{t}, \vec{c}_{u} \in \setsym{R}^{n}$ represents the $filtered$ signal at time $t$. The last step Eq.~\eqref{eq:CTCAttn-TimeConvolution} holds when $\alpha_{u,t} = \frac{1}{C}$ and $\gamma = C$. Since Eq.~\eqref{eq:CTCAttn-TimeConvolution}
is similar to Eq.~\eqref{eq:RNNED-annotate} in structure, $\vec{c}_{u}$ represents a special case context vector with uniform attention weights $\alpha_{u,t} = \frac{1}{C}$, $t \in [u-\tau, \ u+\tau]$. Moreover, $\vec{c}_{u}$ is a result of convolving features $\vec{h}$ with $\vec{W}^{\prime}$ in time. Thus, $\vec{W}^{\prime}$ and $\vec{c}_{u}$ represent \textit{time convolution kernel} and \textit{time convolution feature} respectively.

\subsection{Content Attention (CA) and Hybrid Attention (HA)}
\label{ssec: CTCAttn-attn}
\begin{figure}
\centering
\resizebox{0.70\linewidth}{!}{
\includegraphics[width=\textwidth,height=0.4\textwidth,trim=1mm 4mm 0mm 2mm,clip]{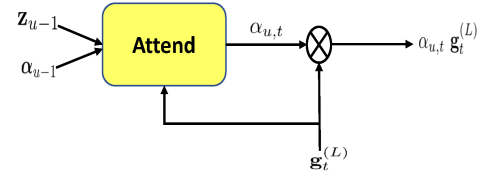}\vspace{-5mm}
}
\caption{Content and hybrid attention.}
\label{fig:CA_HA}\vspace{-4mm}
\end{figure}

To incorporate non-uniform attention in Eq.~\eqref{eq:CTCAttn-TimeConvolution}, we need to compute a non-uniformly distributed $\bm{\alpha}_{u}$ where $\bm{\alpha}_{u} = (\alpha_{u-\tau}, \cdots, \alpha_{u}, \cdots, \alpha_{u+\tau})$ using an attention network similar to Eq.~\eqref{eq:RNNED-attend}. 
However, since there is no explicit decoder like Eq.~\eqref{eq:RNNED-recurrent} in CTC, there is no decoder state $\vec{s}_{u}$. Therefore, we use $\vec{z}_{u}$ instead of $\vec{s}_{u}$. The term $\vec{z}_{u}  \in \setsym{R}^{K}$ is the logit to the softmax and is given by
\begin{align}
\vec{z}_{u} &= \vec{W}_{\text{soft}}\vec{c}_{u} + \vec{b}_{\text{soft}}, \nonumber \\
\vec{p}(\pi_{u}|\vec{x}) &= \text{Softmax}(\vec{z}_{u}), \label{eq:CTCAttn-generate}
\end{align}
where  $\vec{W}_{\text{soft}} \in \setsym{R}^{K \times n}, \vec{b}_{\text{soft}} \in \setsym{R}^{K}$. The term $\vec{p}(\pi_{u}|\vec{x}) = [p(\pi_{u}=1|\vec{x})  \ p(\pi_{u}=2|\vec{x}) \cdots p(\pi_{u}=K|\vec{x})]^{\text{T}}$ is the vector of probabilities of labels in the alignment at time $u$. Thus, Eq.~\eqref{eq:CTCAttn-generate} is similar to the Generate(.) function in Eq.~\eqref{eq:RNNED-generate} but lacks the dependency on $\vec{y}_{u-1}$ and $\vec{s}_{u}$. Consequently, the Attend(.) function in Eq.~\eqref{eq:RNNED-attend} becomes
\begin{align}
\alpha_{u,t} &= \text{Attend}(\vec{z}_{u-1}, \bm{\alpha}_{u-1}, \vec{g}_{t}), \quad t = u-\tau, \cdots, u+\tau \label{eq:CTCAttn-attend}
\end{align}
where $\vec{h}_{t}$ in Eq.~\eqref{eq:RNNED-attend} is replaced with $\vec{g}_{t}$. The Attend(.) function is illustrated in Fig.~\ref{fig:CA_HA} and is simply a single layer neural network with a softmax. A scoring function Score(.), similar to Eq.~\eqref{eq:RNNED-score}, computes the layer activations.
However, here the Score(.) function uses the filtered signal $\vec{g}_{t}$ instead of the raw signal $\vec{h}_{t}$ in Eq.~\eqref{eq:RNNED-score}. Thus, the new Score(.) function becomes
\begin{align}
e_{u,t} &= \text{Score}(\vec{z}_{u-1}, \bm\alpha_{u-1}, \vec{g}_{t}), \quad t = u-\tau, \cdots, u+\tau \label{eq:CTCAttn-score} \\
&= \begin{cases}
\vec{v}^{T}\text{tanh}(\vec{U} \vec{z}_{u-1} + \vec{W} \vec{g}_{t} + \vec{b}), \ \mbox{(content)} \\
\vec{v}^{T}\text{tanh}(\vec{U} \vec{z}_{u-1} + \vec{W} \vec{g}_{t} + \vec{V} \vec{f}_{u} + \vec{b}) \ \mbox{(hybrid)}
\end{cases} \label{eq:CTCAttn-ContentHybrid}
\end{align}
with $\vec{f}_{u}$ a function of $\bm\alpha_{u-1}$ through Eq.~\eqref{eq:RNNED-locfeat}. The content and location information are encoded in $\vec{z}_{u-1}$ and $\bm\alpha_{u-1}$ respectively. Thus, the hybrid function in Eq.~\eqref{eq:CTCAttn-ContentHybrid} includes both content and location information. Scores from Eq.~\eqref{eq:CTCAttn-score} can be normalized using the softmax operation (as in Eq.~\eqref{eq:RNNED-normalizedscore}) to generate non-uniform $\alpha_{u, t}$ for $t \in [u-\tau, \ u+\tau]$. Now, $\bm\alpha_{u}$ can be plugged into Eq.~\eqref{eq:CTCAttn-TimeConvolution}, along with $\vec{g}$ to generate the context vector $\vec{c}_{u}$. This completes the attention network. We found that excluding the scale factor $\gamma$ in Eq.~\eqref{eq:CTCAttn-TimeConvolution}, even for non-uniform attention, was detrimental to the final performance. Therefore, we continue to use $\gamma = C$.

\subsection{Pseudo Language Model (PLM)}
\label{ssec: CTCAttn-LM} 
\begin{figure}
\centering
\resizebox{0.60\linewidth}{!}{
\includegraphics[width=\textwidth,height=0.48\textwidth,trim=1mm 1mm 1mm 0mm,clip]{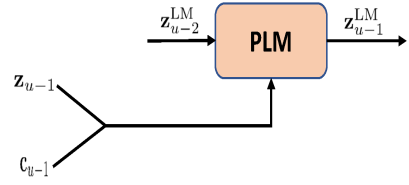}\vspace{-2mm}
}
\caption{Pseudo language model.}
\label{fig:PLM}\vspace{-4mm}
\end{figure}

The performance of the attention model can be improved further by providing more reliable content information from the past. This is possible by introducing another recurrent network, which we refer to as PLM, that can utilize content from several time steps in the past instead of just one. This network, in essence, would learn an LM-like model implicitly. This is illustrated in  Fig.~\ref{fig:PLM}. To build the PLM network, we follow an architecture similar to RNN-LM \cite{Mikolov-RNNLM}. As illustrated in the PLM block of Fig.~\ref{fig:CTCAttn}, the input to the PLM network is computed by stacking the previous output $\vec{z}_{u-1}$ with the context vector $\vec{c}_{u-1}$ and feeding it to a recurrent function $\mathcal{H}(.)$. The output of $\mathcal{H}(.)$ is $\vec{z}^{\text{LM}}_{u-1}$ which, instead of $\vec{z}_{u-1}$, is fed to the Attend(.) block in Eq.~\eqref{eq:CTCAttn-attend}. This is represented as
\begin{align}
\vec{z}^{\text{LM}}_{u-1} &= \mathcal{H}(\vec{x}_{u-1}, \vec{z}^{\text{LM}}_{u-2}), \quad
\vec{x}_{u-1} = 
\begin{bmatrix}
\vec{z}_{u-1} \\
\vec{c}_{u-1}
\end{bmatrix}, \label{eq:CTCAttnLM-LSTM} \\
\alpha_{u,t} &= \text{Attend}(\vec{z}^{\text{LM}}_{u-1}, \bm{\alpha}_{u-1}, \vec{g}_{t}), \quad t = u-\tau, \cdots, u+\tau .  \label{eq:CTCAttn-attendLM}
\end{align}
We model $\mathcal{H}(.)$ using a single layer long short-term memory (LSTM) unit \cite{Hochreiter1997long} with $n$ memory cells and input and output dimensions set to $K + n$ (since $\vec{x}_{u-1} \in \setsym{R}^{K+n}$) and $n$ (since $\vec{z}^{\text{LM}}_{u-1} \in \setsym{R}^{n}$) respectively. Notice that $\vec{z}^{\text{LM}}_{u-1}$ encodes the content of a pseudo LM rather than a true LM since CTC outputs are interspersed with blank symbols by design. Also, $\vec{z}^{\text{LM}}_{u-1}$ is a real-valued vector instead of a one-hot vector. Hence, the PLM is not a true LM.

\subsection{Component Attention (COMA)}
\label{ssec: CTCAttn-Comp}
In the previous sections, $\alpha_{u,t}$ is a scalar term weighting the contribution of the entire $n$-dimensional vector $\vec{g}_{t}$ to generate the output $\vec{p}(\pi_{t}|\vec{x})$. This means all $n$ components (or dimensions) of the vector $\vec{g}_{t}$ are weighted by the same scalar $\alpha_{u,t}$. In this section, we consider weighting each component (dimension) of $\vec{g}_{t}$ using a separate weight. Therefore, we need an $n$-dimensional weight vector $\bm\alpha_{u,t} \in \setsym{U}^{n}$ instead of the scalar $\alpha_{u,t} \in \setsym{U}$. The vector $\bm\alpha_{u,t}$ can be generated as follows. First, compute an $n$-dimensional score $\vec{e}_{u, t}$ for each $t$. This is easily achieved using the Score(.) function in Eq.~\eqref{eq:CTCAttn-ContentHybrid} but without taking the inner product with $\vec{v}$. For example, in the case of hybrid, the scoring function becomes
\begin{align}
\hspace{-8pt}\vec{e}_{u, t}&=\text{tanh}(\vec{U} \vec{z}_{u-1} + \vec{W} \vec{g}_{t}  + \vec{V} \vec{f}_{u} + \vec{b}), \ t = u-\tau, \cdots, u+\tau. \label{eq:CTCAttn-Comp-score}
\end{align}
Now, we have $C$ column vectors $[\vec{e}_{u, u-\tau}, \cdots, \vec{e}_{u, u+\tau}]$ where each vector is of dimension $n$. Stacking them column-wise, we have an $n \times C $ scoring matrix $\vec{E}$
\vspace{-2mm}
\begin{align}
\vec{E} &= 
\begin{bmatrix} 
 \vertbar            & \vertbar             &           & \vertbar \\ 
 \vec{e}_{u, u-\tau}   & \vec{e}_{u, u - \tau + 1}    & \ldots    & \vec{e}_{u, u+\tau} \\
 \vertbar            & \vertbar             &           & \vertbar
\end{bmatrix}_{n \times C}.
\label{eq:coma-score-splice}
\end{align}
Let $e_{u, t}(j) \in (-1,1)$ be the $j^{\text{th}}$ component of the vector $\vec{e}_{u, t}$. To compute $\alpha_{u, t}(j)$ from $e_{u, t}(j)$, we normalize $e_{u, t}(j)$ across $t$ (columns) keeping  $j$ (row) fixed. Thus, $\alpha_{u, t}(j)$ is computed as
\begin{align}
\alpha_{u, t}(j) &= \frac{\text{exp}(e_{u, t}(j))}{\sum_{t^{\prime}=u-\tau}^{u+\tau} \text{exp}(e_{u, t^{\prime}}(j))}, \quad j=1,\cdots,n. \label{eq:CTCAttn-Comp-scoresoftmax}
\end{align}
Since $\text{exp}(.)$ and $\text{tanh}(.)$ are both one-to-one functions, their composition is also one-to-one. Thus, there is a one-to-one correspondence between  the input $g_{t}(j)$ and output $\alpha_{u, t}(j)$ through the composite function. Consequently, $\alpha_{u, t}(j)$ can be interpreted as the amount of contribution of $g_{t}(j)$ in computing $c_{u}(j)$. Now, from Eq.~\eqref{eq:CTCAttn-Comp-scoresoftmax}, we know the values of the vectors $\bm\alpha_{u,t}$, $t \in [u-\tau, \ u+\tau]$. Hence, under the COMA formulation, the context vector $\vec{c}_{u}$ can be computed from $\bm\alpha_{u,t}$ and $\vec{g}_{t}$ using
\begin{align}
\vec{c}_{u} &= \text{Annotate}(\bm\alpha_{u}, \vec{g}, \gamma) = \gamma \sum_{t=u-\tau}^{u+\tau} \bm\alpha_{u,t} \odot \vec{g}_{t}, \label{eq:CTCAttn-Comp-annotate}
\end{align}
where $\odot$ is the Hadamard product. One attractive feature of the COMA formulation is that it does not introduce any additional training parameters.

Finally, we highlight the differences between the attention mechanism in this work and in \cite{prabhavalkar2017comparison}. First, we apply attention across time (past, present, future) on the time convolution features extracted from the final layer of the recurrent network (encoder). Moreover, we attend only to a small context window. In contrast, \cite{prabhavalkar2017comparison} attends to the entire output sequence of the encoder in addition to the state of the decoder. There is no time convolution applied on the encoder sequence either. Second, to improve attention modeling, we make use of the logit from the previous time $\vec{z}_{u-1}$ (or  $\vec{z}^{\text{LM}}_{u-1}$) as an additional input to our attention block. The attention mechanism in \cite{prabhavalkar2017comparison} does not make use of logit due to the presence of an explicit decoder. Finally, our COMA formulation yields additional gains without introducing any additional training parameters. There is no such formulation in \cite{prabhavalkar2017comparison}.

\section{Self-Attention CTC}
\label{sec: SelfAttnCTC}
In this section, we investigate another attention-based paradigm known as Self-Attention (SA) \cite{vaswani2017selfattention} in the context of CTC training. There are some key differences in the way the attention weights are computed between SA-CTC and Attention CTC (Section~\ref{sec: CTCAttn}). In Attention CTC, the attention weights are computed using the hidden features and the output prediction from the previous time step ($\vec{z}_{u-1}$). This is evident from the scoring function in Eq.~\eqref{eq:CTCAttn-score}. In contrast, in SA-CTC, the weights are computed from the hidden features only. It does not use any past output predictions. 
Another difference is that the attention weights are computed using additive operations in Attention CTC whereas multiplicative operations (inner products) are used in SA-CTC. Moreover, matrix-vector multiplications used in Attention CTC are computationally slower than performing inner products in SA-CTC. 

We highlight only the most important steps in the formulation of SA-CTC. First the hidden features are converted into input projections using the projection matrix $\vec{W}_{p}$ as
\begin{align}
\vec{b}_{t} &= \vec{W}_{p} \vec{h}_{t}, \quad t = u-\tau, \cdots, u+\tau
\end{align}
where $u$ denotes the current time step. The inputs to the attention block of SA-CTC consists of three kinds of vectors - keys, values, and a query. These are derived using
\begin{align}
\vec{q}_{t} &= \vec{Q} \vec{b}_{t}, \quad t = u, \\
\vec{k}_{t} &= \vec{K} \vec{b}_{t}, \quad t = u-\tau, \cdots, u+\tau, \\
\vec{v}_{t} &= \vec{V} \vec{b}_{t}, \quad t = u-\tau, \cdots, u+\tau,
\end{align}
where $\vec{Q}, \vec{K}, \vec{V}$ are the query, key, and value matrices respectively. Here, the dimensions of $\vec{q}_{t}, \vec{k}_{t}, \vec{v}_{t}$ are $d_{k}, d_{k}, d_{v}$ respectively. Note that while there is a single query vector corresponding to the current time step $u$, there are multiple key and value vectors corresponding to the context window $[u-\tau, u+\tau]$. 

Following this, scores are evaluated between the query and the keys by taking their dot products and scaling them with $\frac{1}{\sqrt{d_{k}}}$. This is given by
\begin{align}
e_{u, t} &=  \frac{\vec{q}^{T}_{u} \vec{k}_{t}}{\sqrt{d_{k}}}, \quad t = u-\tau, \cdots, u+\tau.
\end{align}
The scores reflect the correlation between the current input and the neighboring inputs. These scores are then converted into probabilities (attention weights) using the softmax operation. Linear combination of the value vectors using these attention weights generates a context vector $\vec{c}_{u}$ as follows:
\begin{align}
\alpha_{u,t} &= \frac{\text{exp}(e_{u, t})}{\sum_{t^{\prime}=u-\tau}^{u+\tau} \text{exp}(e_{u, t^{\prime}})},  \quad t = u-\tau, \cdots, u+\tau \\
\vec{c}_{u} &= \sum_{t=u-\tau}^{t=u+\tau} \alpha_{u,t} \vec{v}_{t}.
\end{align}

This is followed by a residual connection \cite{he2017deepresidual} and layer normalization, i.e., $\text{LayerNorm}(\vec{c}_{u} + \vec{b}_{u})$. The output of this is fed to a single layer feed-forward network which is followed by another round of residual connection and layer normalization. 
This is the uni-head attention architecture of SA-CTC since it computes a scalar weight $\alpha_{t}$ for the entire value vector $\vec{v}_{t}$. This can be easily extended to multi-head attention where $\vec{v}_{t}$ is fragmented into smaller sub-vectors and each sub-vector is weighted using a distinct scalar weight. For more details on SA architecture, readers may refer \cite{vaswani2017selfattention}.

\section{Hybrid CTC}
\label{sec: hybCTC}
In this and the next section, our primary motivation is to mitigate the OOV issue of the A2W model as mentioned in Section \ref{sec: Intro}.

\begin{figure}
\centering
\resizebox{0.90\linewidth}{!}{
\includegraphics[width=0.5\textwidth,height=0.5\textwidth,trim=8mm 0mm 2mm 2mm,clip]{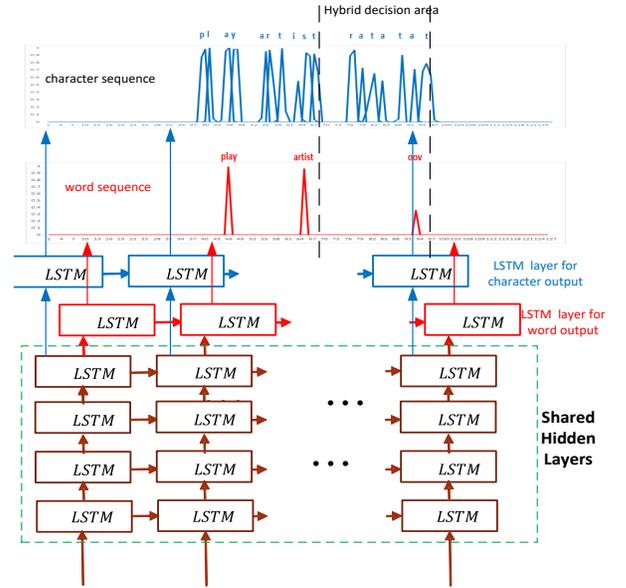}
}
\caption{An example of how the Hybrid CTC solves the OOV issue of the word CTC. The words ``play", ``artist", ``OOV" are obtained from the word CTC. The words ``play", ``artist", ``ratatat" are obtained from the letter CTC. Hence, the final output of Hybrid CTC is ``play, artist, ratatat" with the first two words obtained from the word CTC and the last word obtained from the letter CTC.}
\label{fig:hybCTC}
\end{figure}

First, we describe the Hybrid CTC network. The Hybrid CTC network uses a word CTC as the primary task and a letter CTC as the auxiliary task in an MTL framework. The output units of the word CTC correspond to frequently used words and an OOV token. Infrequent words in the training set are lumped together and tagged as OOV. Given an input sequence of features, the word and letter CTCs emit a word and letter sequence respectively. If the word sequence contains a list of frequent words, then the letter sequence from the letter CTC is completely ignored. However, if the word sequence contains the OOV token, the letter CTC is consulted at the segment that generated the OOV token. In the consultation process, the letter sequence from the letter CTC is merged to form a word.  Finally, this newly constructed word from the letter CTC is used to replace the OOV token. Since the word CTC and letter CTC are time synchronized through the shared hidden layers of the MTL network, it is possible to find a correspondence between the outputs of the two CTCs. An illustration of this method is shown in Fig.~\ref{fig:hybCTC}. Here, the word CTC generates the sequence ``\textit{play artist OOV}". The word sequence generated after merging the letters from the letter CTC is ``\textit{play artist ratatat}". Since the segment containing ``\textit{ratatat}" from the letter CTC has the most time overlap with the segment containing ``\textit{OOV}" from the word CTC, the OOV token is replaced with ``\textit{ratatat}". Thus, the final output of the Hybrid CTC is ``\textit{play artist ratatat}".

The detailed steps for building the Hybrid CTC model are described as follows:
\begin{itemize}
	\item Build an LSTM-CTC model of $L$ layers with its output units mapped to frequently occurring words in the training corpus. Map all the remaining infrequent words (occurring less than $N$ times) as the OOV token. Thus, the output units in this LSTM-CTC model correspond to (a) the frequent words, (b) the OOV token, and (c) blank and silence (two additional tokens).
	\item Freeze the bottom $L-1$ hidden layers of the word-CTC, add one LSTM hidden layer and one softmax layer to build a new LSTM-CTC model with letters as its output units. 
	\item During testing, generate the word output sequence using greedy decoding. If the output word sequence contains an OOV token, replace the OOV token with the word generated from the letter CTC that has the largest time overlap with the OOV token.
\end{itemize}

\section{Mixed-unit CTC}
\label{sec: multimixCTC}

In this section, we briefly explain multi-letter CTC and compare the past implementation of multi-letter CTC with ours. Based on this foundation, we then explain our proposed Mixed-unit CTC.


Although single-letter units in CTCs perform well, they are prone to high degree of variability across training examples due to their short temporal context. As we will see later in Table \ref{Tab:WER_CTC_multiletter}, multi-letter units tend to perform better than single-letter units since they have low degree of variability by capturing context information and thereby offer more stability during training. Improving letter CTCs can help improve the accuracy of word CTCs. For example, a stronger letter CTC can lower the WER of the Hybrid CTC since the OOV token may be replaced by more precise words generated by the letter CTC.
 
Gram CTC \cite{liu2017gram} and multi-phone CTC \cite{siohan2017ctc} are multi-letter CTCs based on letters and phonemes respectively. They allow variable number of letters (or grams) and phonemes to be output at each time step. The size of the units in gram CTC and multi-phone CTC are learned automatically with the modified forward-backward algorithm accounting for all decompositions. However, in the test phase, their decoding procedure is more complex than the simple greedy decoding procedure used in single-letter CTC models. To reduce the decoding complexity, the authors in \cite{Chen2017PhoneSynchronous} proposed phone synchronous decoding. In contrast, we offer a facile implementation of our multi-letter CTC. We simply decompose every word (which includes both frequent and OOV words) into a sequence of one or more letter units. Examples are shown in the first three rows of Table~\ref{Tab:units} where each word, frequent or OOV, is decomposed into single-letter or double-letter or triple-letter units. The advantages of doing this are three-fold. First, our decomposition is straightforward. Second, it does not change the CTC forward-backward algorithm. Finally, during the test phase, our method is able to retain the same greedy decoding procedure used in single-letter CTC models.

\begin{table}[!t]
\centering
\caption{Examples of how words are decomposed into different output units. ``Newyork" is a frequent word while ``newyorkabc" is an OOV (infrequent word).} 
\scalebox{0.75}{
\begin{tabular}{l|c c}
   \hline

		Decomposition Type					&newyork      &newyorkabc     \\\hline
All words $\rightarrow$ single-letter		&n e w y o r k      &n e w y o r k a b c  \\
All words $\rightarrow$ double-letter       &ne wy or k      &ne wy or ka bc   \\
All words $\rightarrow$ triple-letter       &new yor k      &new yor kab c   \\ 
All words $\rightarrow$ word				&newyork      &OOV	\\
OOVs only $\rightarrow$ single-letter		&newyork      &n e w y o r k a b c \\
OOVs only $\rightarrow$ word+single-letter 	&newyork 	&newyork a b c   \\
OOVs only $\rightarrow$ word+triple-letter 	&newyork 	&newyork abc   \\ \hline
\end{tabular}
}
\label{Tab:units}
\end{table}

\begin{figure}[!t]
\centering
\resizebox{0.90\linewidth}{!}{
\includegraphics[width=0.5\textwidth,height=0.5\textwidth,trim=8mm 0mm 2mm 2mm,clip]{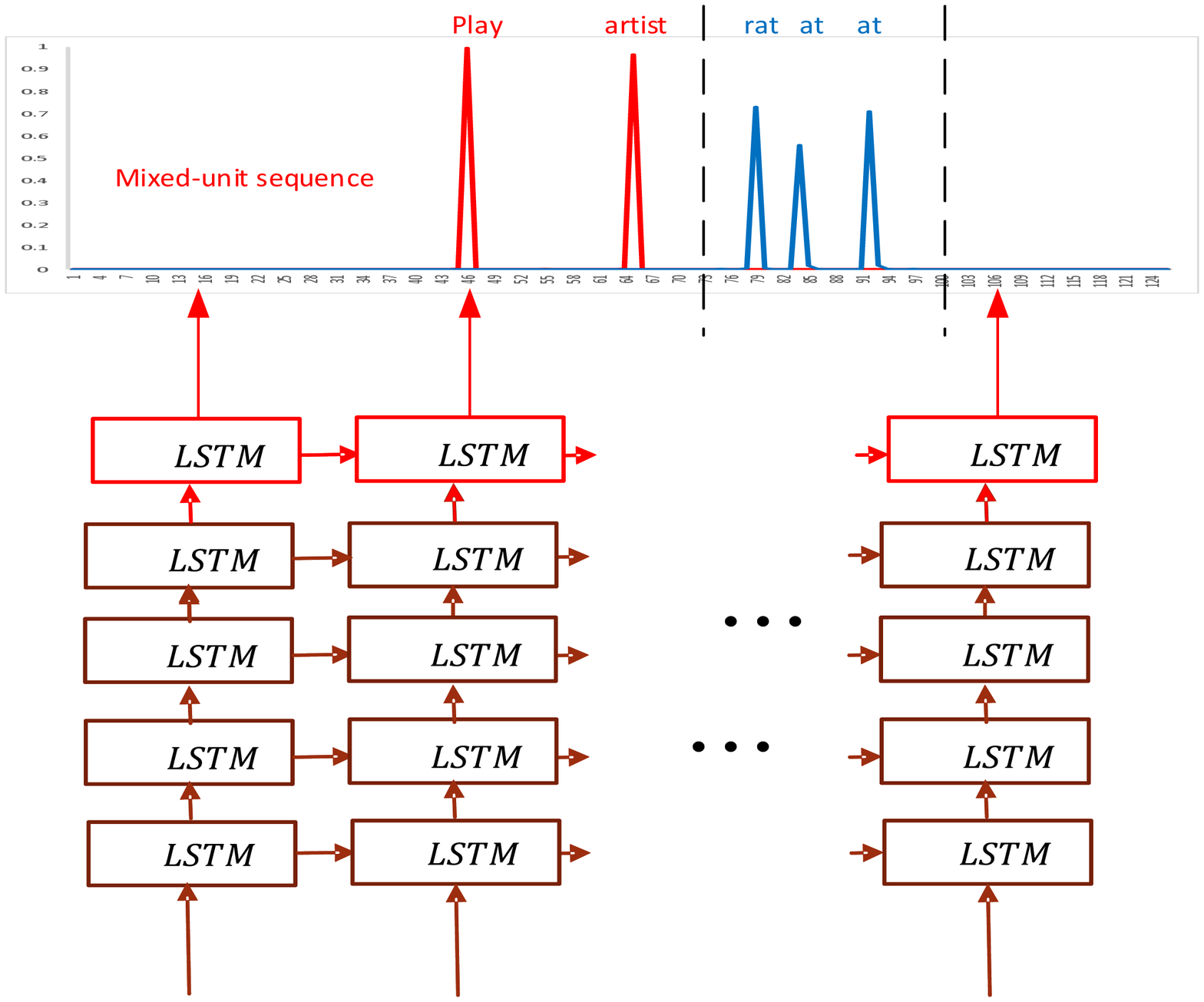}
}
\caption{An example of how a single CTC trained with a mixture of words and letters solves the OOV issue. The final output of this CTC is ``play, artist, rat at at".}
\label{fig:mixCTC}\vspace{-2mm}
\end{figure}

In Hybrid CTC, the shared-hidden-layer constraint is used to aid the time synchronization of word outputs between the word and letter CTCs. However, the blank symbol dominates most of the frames. The unit boundaries from CTC is also notorious for being arbitrary. Therefore, time synchronization may not be very reliable with the two CTCs running in parallel. A direct solution is to forgo the MTL framework and train a single CTC model comprising of a mixture of frequent words and letters. The letters arise as a result of decomposing the infrequent words in the training set into letters before CTC training begins. The working of this CTC is illustrated in Fig.~\ref{fig:mixCTC}. If the word is a frequent word, then we just keep it in the output token list. If it is an OOV, then we decompose it into a letter sequence. As shown in the fifth row of Table~\ref{Tab:units}, the OOV ``newyorkabc" is decomposed into ``n e w y o r k a b c'' for single letter decompositions. However, the word ``newyork" is not decomposed any further because it is a frequent word. Therefore, the output units of the CTC are both words (for frequent words) and letters (for OOVs). 

However, we note that artificially decomposing OOVs into sequences of single-letters only may confuse CTC training because the network output modeling units are frequent words and letters.  To solve such a potential issue, we decompose the OOVs into a combination of frequent words and letters. We refer to this combination as \textit{mixed units}. For example, in the last two rows of Table~\ref{Tab:units}, the OOV ``newyorkabc" is decomposed into ``newyork a b c'' if we use words and single-letter units or ``newyork abc'' if we use words and triple-letter units. In addition, for mixed units, we use ``\$'' to separate each word in the sentence.  For example, the sentence ``have you been to newyorkabc'' is decomposed into ``\$ have \$ you \$ been \$ to \$ newyork abc \$''. 
The ``\$'' symbol acts as a word separator (like the space symbol) and is essential for finding word boundaries of the mixed-units. During training, since the OOVs are decomposed into mixed units, there is no ``OOV" output node in the Mixed-unit CTC model. Consequently, during testing, the model emits mixed units instead of ``OOV" while still emitting frequent words.

\section{Experiments}
\label{sec: Expts}

In this section, we compare the performance of the proposed CTCs with the baseline CTC. We evaluated the proposed methods using Microsoft's Cortana voice assistant task.  The training and test sets consist of approximately 3400 hours ($\sim$ 3.3 million utterances) and 6 hours ($\sim$ 5600 utterances) of audio spoken in American English respectively.

All CTC models were trained using either unidirectional LSTMs (ULSTM) or bidirectional LSTMs (BLSTM). The ULSTM is a 5-layer LSTM with 1024 memory cells in each layer. Similarly, the BLSTM is a 6-layer LSTM with 512 memory cells in each direction (therefore resulting in 1024 output dimensions when combining outputs from both directions). The cell outputs are linearly projected to 512 dimensions. The base feature vector is a 80-dimensional vector containing log filterbank energies computed every 10 ms. Eight frames of base features were stacked together ($m = 80 \times 8 = 640$) as the input to the unidirectional CTC, while three frames were stacked together ($m =  80 \times 3 = 240$) as the input to the bidirectional CTC. The skip rate for both unidirectional and bidirectional CTCs was three frames as in \cite{sak2015fast}. The dimension $n$ of vectors $\vec{h}_{t}, \vec{g}_{t}, \vec{c}_{u}$ was set to 512. For decoding, the greedy decoding procedure (no complex beam search decoder or external LM) was used. This makes our E2E ASR systems purely all-neural.

We focus on letter CTC first and then move on to word CTC. This is because improvements in the letter CTC increase the accuracy of the word CTC especially when encountering an OOV word during test time. Thus, we evaluated the performance of Attention CTC (Section~\ref{sec: CTCAttn}), SA-CTC (Section~\ref{sec: SelfAttnCTC}), and multi-letter CTC using letter units. Then, we evaluated the performance of our proposed Hybrid CTC (Section~\ref{sec: hybCTC}) and Mixed-unit CTC (Section~\ref{sec: multimixCTC}) using both word and letter units.

\subsection{Experiments With Letter-Based CTCs}
\label{ssec: Expts1}

We experimented with different sizes of letter units. The sizes are represented by the cardinality $K$ of the label set (defined in Section~\ref{ssec: CTC}). For single-letter units, $K$ was set to 30. This corresponds to 26 English letters [a-z], ', *, \$, and a blank symbol. For double and triple-letter units, $K$ was set to 763 and 8939 respectively covering all double-letter and triple-letter occurrences in the training set. 

\subsubsection{Attention CTC (Section~\ref{sec: CTCAttn})}
\label{sssec: exp_CTCAttn_letter}
In the first set of experiments, we evaluated Vanilla CTC \cite{Graves-CTCFirst} and the proposed Attention CTC models trained using our 5-layer ULSTM with single-letter units. We experimented with $\tau = 4$ (length of one-sided attention window, defined in Section~\ref{ssec: CTCAttn-conv}) considering the training efficiency with this setting. The results are tabulated in the second column of Table \ref{Tab:WER_CTCAttn_ULSTM_BLSTM_letter}. The top row presents the WER for Vanilla CTC. All subsequent rows under ``Attention CTC" present the WER for the proposed Attention CTC models when attention modeling capabilities were gradually added in a stage-wise fashion. The best proposed model is in the last row. It includes component attention (COMA) along with all the other enhancements above it (i.e., TC, HA, PLM). It may be recalled, from  Eq.~\eqref{eq:CTCAttn-ContentHybrid}, that hybrid attention (HA) is a combination of both content and location attention. The best proposed model outperformed Vanilla CTC by 22.72\% relative. We found that the gains are marginal when going from CA to HA. Our conjecture is that the benefits of adding location information in HA could become more pronounced with smaller frame sizes and larger attention windows. However, smaller frame sizes lead to an exponential increase in the number of CTC paths resulting in instability during CTC training. 

\begin{table}[!t]
\centering
\caption{WERs of letter-based Vanilla CTC \cite{Graves-CTCFirst} and Attention CTC for $\tau = 4$
trained with 5-layer ULSTM or 6-layer BLTSM and single-letter output units. Relative WER reduction and the number of model parameters are in parentheses. TC = Time Convolution, CA = Content Attention, HA = Hybrid Attention, PLM = Pseudo Language Model, COMA = Component Attention, M = million.}
\scalebox{0.75}{
\begin{tabular}{l|c c}
   \hline
E2E Model  & \multicolumn{2}{c}{WER (\%)} \\ \cline{2-3} 
           & ULSTM & BLSTM \\
\hline
Vanilla CTC	      &24.03 (0.00, 24.12 M)   & 17.84 (0.00, 35.13 M)  \\ \hline
Attention CTC     &   			  &		   \\
+TC (Sec \ref{ssec: CTCAttn-conv})   &21.89 (08.91, 26.48 M) &17.67 (0.95, 37.49 M) \\
+CA (Sec \ref{ssec: CTCAttn-attn})   &20.45 (14.90, 26.74 M) &16.13 (09.59, 37.75 M) \\
+HA (Sec \ref{ssec: CTCAttn-attn})   &20.27 (15.65, 27.00 M) &16.01 (10.26, 38.01 M) \\
+PLM (Sec \ref{ssec: CTCAttn-LM})    &19.78 (17.69, 29.62 M) &15.34 (14.01, 40.63 M) \\
+COMA (Sec \ref{ssec: CTCAttn-Comp}) &\bf{18.57 (22.72, 29.62 M)} &\bf{14.47 (18.89, 40.63 M)} \\ \hline
\end{tabular}
}
\label{Tab:WER_CTCAttn_ULSTM_BLSTM_letter}
\end{table}

Next we evaluated Attention CTC models trained with our 6-layer BLSTM. The results are tabulated in the third column of Table~\ref{Tab:WER_CTCAttn_ULSTM_BLSTM_letter}. Similar to the unidirectional case, the best proposed model outperformed Vanilla CTC by 18.89\% relative. This shows that the proposed Attention CTC models continue to perform well even with stronger baselines like BLSTMs.

As an additional experiment, we compared RNN-T models trained with 5-layer ULSTM or 6-layer BLSTM transcription networks along with 1-layer ULSTM prediction network and 30 letters as output units. The transcription networks have the same structure as our baseline CTC models. We observed 21.07\% and 16.96\% WER for ULSTM and BLSTM transcription networks respectively. While this outperforms the baseline CTC error rates reported in Table \ref{Tab:WER_CTCAttn_ULSTM_BLSTM_letter}, it could not outperform our final Attention CTC model (last row in Table \ref{Tab:WER_CTCAttn_ULSTM_BLSTM_letter}).

\subsubsection{Self-Attention CTC (Section~\ref{sec: SelfAttnCTC})}
\label{sssec: exp_SACTC_letter}
In the next set of experiments, we evaluated the performance of SA-CTC models using our ULSTM and BLSTM with attention window size $\tau = 4$. We used 1024-dimensional vectors for both key/query and value vectors. Thus, $d_{k} = 1024,  d_{v} = 1024$. This is in accordance with the number of memory cells used in Attention CTC. We experimented with other dimensions but they performed worse. Furthermore, we experimented with both single and multi-head attention (4 and 8 heads). The results are tabulated in Table~\ref{Tab:WER_SelfAttnCTC_letter}. SA-CTC with 8 heads performed the best for each case. The relative WERR over Vanilla CTC are 21.56\% and 16.59\% using ULSTM and BLSTM respectively. Comparing the best models from Attention CTC and SA-CTC, we find that Attention CTC performed slightly better than SA-CTC by about 1.2\%  (22.72-21.56) and 2.3\% (18.89-16.59) for ULSTM and BLSTM respectively.

\begin{table}[!t]
\centering
\caption{WERs of letter-based Vanilla CTC \cite{Graves-CTCFirst} and SA-CTC (1, 4, 8 heads) for $\tau = 4$ trained with ULSTM/BLSTM and single-letter units.  Relative WER reduction and the number of model parameters are in parentheses. M = million.}
\scalebox{0.75}{
\begin{tabular}{l|c c}
   \hline
E2E Model & \multicolumn{2}{c}{WER (\%)} \\ \cline{2-3} 
           & ULSTM & BLSTM \\ \cline{2-3}
            \hline
Vanilla CTC      &24.03 (0.00, 24.12 M)  &17.84 (0.00, 35.13 M)  \\ \hline
SA-CTC (1 head)  &20.06 (16.52, 30.70 M) &15.69 (12.05, 41.91 M) \\
SA-CTC (4 heads) &18.90 (21.35, 30.71 M)	&14.98 (16.03, 41.92 M) \\
SA-CTC (8 heads) &\bf{18.85	(21.56, 30.72 M)}	&\bf{14.88 (16.59, 41.93 M)} \\
 \hline
\end{tabular}
}
\label{Tab:WER_SelfAttnCTC_letter}
\end{table}

\subsubsection{Multi-letter CTC}
\label{sssec: exp_letter}
In the next set of experiments, we evaluated the performance of various CTC models trained using our 6-layer BLSTM with multi-letter units as outputs. We evaluated three kinds of CTC models: Vanilla CTC, Attention CTC, and Attention CTC sharing 5 hidden layers with a word CTC. 
In the third CTC model, we applied attention only to the letter CTC.

As shown in the third column of Table~\ref{Tab:WER_CTC_multiletter}, the WER of Vanilla CTC drops significantly when the output units become larger (and hence more stable). The letter CTC using triple-letter units achieved 13.28\% WER which is a relative WERR of 25.56\% compared to the letter CTC using single-letter units.

As shown in the fourth column of Table~\ref{Tab:WER_CTC_multiletter}, Attention CTC improves hugely over the Vanilla CTC. It achieves about 18.89\%, 20.88\%, and 14.46\% relative WERR over Vanilla CTC using single-letter, double-letter, and triple-letter units respectively. 

In the last column of Table~\ref{Tab:WER_CTC_multiletter}, the shared Attention CTC performed better than the Vanilla CTC but worse than its non-sharing counterpart. This indicates one shortcoming of the shared Attention CTC -- it sacrifices the accuracy of the letter CTC because of the shared-hidden-layer constraint with the word CTC.
 

\begin{table}[!t]
\centering
\caption{WERs of letter-based CTC models, trained using 6-layer BLSTMs, with Multi-letter output units. Three structures are evaluated: Vanilla CTC \cite{Graves-CTCFirst}, Attention CTC, and Attention CTC sharing 5 hidden layers with word CTC. One-sided attention window size ($\tau$) set to 4.}
\scalebox{0.90}{
\begin{tabular}{l|c|c c c}
   \hline
E2E Model  & Total &\multicolumn{3}{c}{WER (\%)} \\ \cline{3-5}
       	   & Units &Vanilla  &Attention  &Attention + \\
		   & &         &     &5 layers sharing \\\hline
single-letter	 &30       &17.84 &14.47   &16.74   \\
double-letter    &763      &15.37 &12.16   &14.00  \\
triple-letter    &8939     &\bf{13.28} &\bf{11.36}   &\bf{12.81}  \\ \hline
\end{tabular}
}
\label{Tab:WER_CTC_multiletter}
\end{table}


\subsection{Experiments With Word-Based CTCs}
\label{ssec: Expts2}
In this section, we evaluate the performance of the Hybrid CTC (Section~\ref{sec: hybCTC}) and the Mixed-unit CTC (Section~\ref{sec: multimixCTC}) using both words and letters as targets. We refer to these CTCs as word CTCs since a majority of the output nodes in these CTCs directly correspond to words. We are primarily interested in recognizing the OOVs as accurately as possible while also boosting the accuracy of recognizing non-OOVs. All attention models in this section are based on Attention CTC (Section \ref{sec: CTCAttn}) instead of SA-CTC (Section \ref{sec: SelfAttnCTC}) owing to the superior results of the former (Section \ref{sssec: exp_SACTC_letter}).

Our Vanilla CTC \cite{Graves-CTCFirst} is a 6-layer BLSTM with approximately 27k output nodes consisting of frequent words and the OOV token. We defined frequent words as those which occurred at least 10 times in the training corpus. All the remaining words were tagged as OOV. This is the mapping scheme described in the fourth row of Table \ref{Tab:units}. Thus, within the family of word CTCs, the Vanilla CTC is a CTC with 6-layer BLSTM whose output units model words and the OOV token. The Vanilla CTC achieved 9.84\% WER (Table~\ref{Tab:WER_HybCTC_word}) among which the OOVs contributed to 1.87\% WER.


\begin{table}[!t]
\centering
\caption{WERs of word-based Vanilla CTC \cite{Graves-CTCFirst} and Hybrid CTC models trained with 6-layer BLSTMs. Hybrid CTC uses a word CTC and an Attention CTC outputting multi-letter units. Attention models are based on Section \ref{sec: CTCAttn}.}
\scalebox{0.92}{
\begin{tabular}{l|c}
   \hline
E2E Model   & \multicolumn{1}{c}{WER (\%)} \\
 \hline
Vanilla CTC   			       &9.84       \\ \hline
Hybrid CTC: word + double-letter Attention CTC            &9.66     \\
Hybrid CTC: word + triple-letter Attention CTC              &9.66     \\
 \hline
\end{tabular}
}
\label{Tab:WER_HybCTC_word}
\end{table}

\subsubsection{Hybrid CTC (Section~\ref{sec: hybCTC})}
\label{sssec: exp_hybrid}

Our Hybrid CTC model has both word and letter CTCs operating in parallel in an MTL framework. They share 5 hidden BLSTM layers. An additional LSTM layer was added for each task (word and letter CTC) and fine tuned. Thus, the underlying structure of Hybrid CTC is still a 6-layer BLSTM which has the same number of hidden layers as that of the Vanilla CTC. Results are tabulated in Table~\ref{Tab:WER_HybCTC_word}. Both hybrid models achieved 9.66\% WER which is a marginal improvement over the Vanilla CTC. Several factors contribute to such a small improvement. First, the shared-hidden-layer constraint degrades the performance of the letter CTC, potentially affecting the final hybrid system performance. Second, although the shared-hidden-layer constraint helps to synchronize the word outputs from the word and letter CTC, we still observed that the time synchronization can fail at times. In such cases, the OOV token was replaced with its neighboring word because of word segment misalignments. Because of these factors, the triple-letter CTC did not improve over the double-letter CTC.

\subsubsection{Mixed-unit CTC (Section~\ref{sec: multimixCTC})}
\label{sssec: exp_mix}

\begin{table}[!t]
\centering
\caption{WERs of word-based CTCs with different kinds of output units. All CTC models were trained with 6-layer BLSTMs. Attention models are based on Section \ref{sec: CTCAttn}.}
\scalebox{0.90}{
\begin{tabular}{l|c}
   \hline
E2E Model & \multicolumn{1}{c}{WER (\%)} \\
 \hline
Vanilla CTC					    &9.84       \\ \hline
Mixed (OOV $\rightarrow$ single-letter)  &20.10 \\ 
Mixed (OOV $\rightarrow$ word + single-letter)               &10.17 \\
WPM  														 &9.73  \\
Mixed (OOV $\rightarrow$ word + double-letter)               &9.58  \\
Mixed (OOV $\rightarrow$ word + triple-letter)               &9.32  \\
Mixed (OOV $\rightarrow$  word + triple-letter) + Attention  &8.65  \\
 \hline
\end{tabular}
}
\label{Tab:WER_mixCTC_word}
\end{table}

\begin{table}[!t]
\centering
\caption{Summary of WERs of CD phoneme CTC, Vanilla CTC \cite{Graves-CTCFirst}, and Mixed-unit CTC + Attention. All CTC models were trained with 6-layer BLSTMs. Attention models are based on Section \ref{sec: CTCAttn}.}
\scalebox{0.97}{
\begin{tabular}{l|c c}
   \hline
Model & LM & WER(\%) \\
 \hline
1. Conventional: CD phoneme CTC  & \checkmark & 9.28 \\
2. E2E: Vanilla CTC & \ding{55}   & 9.84       \\
3. E2E: Mixed-unit + Attention CTC & \ding{55} & 8.65 \\
 \hline \hline 
 & \#3 vs \#1 &  \#3 vs \#2 \\ \hline
Relative WERR & 6.79\% & 12.09\% \\ \hline
\end{tabular}
}
\label{Tab:WER_summary_wordCTC}
\end{table}

In the next set of experiments, we compared the performance of CTCs by changing their output units to mixed-letter units or wordpieces \cite{Senrich2016NMT, Wu2016WordPiece}.  Wordpieces are commonly occurring sub-word units that can be merged to form whole words. Similar to mixed-units, wordpieces offer the flexibility to generate open-vocabulary words. Previous studies \cite{chan2016latent,rao2017exploring} have explored using wordpieces. To build a wordpiece model (WPM), each word in a training corpus is first segmented into a sequence of individual characters and an end-of-word symbol. Following this, the most frequently occurring character pair is merged to form a new symbol or wordpiece. This process is iterated until a predefined number of wordpieces are generated. The outcome of this is that the corpus is now redefined using those wordpieces which result in minimal number of whole word segmentations. However, our approach of building mixed-units is different from building wordpieces since we decompose \textit{only} OOVs while still retaining the high frequency words as whole word units.



Results are tabulated in Table~\ref{Tab:WER_mixCTC_word}. As before, the Vanilla CTC achieved a WER of 9.84\%. In the next experiment, we decomposed only the OOVs in the training set into single-letters. Thus, the output nodes consist of both single-letters and 27k frequent words. There indeed is no such clear boundary of decomposition with 2 distinct sets of basic units.
As mentioned in Section~\ref{sec: multimixCTC}, having a mixture of word and single-letters confuses CTC training as the network does not know why the frequent words cannot be decomposed into letters. Therefore, this model achieved 20.10\% WER which is far worse than Vanilla CTC. Analyzing the posterior spikes of this model, we observed that the word spikes and letter spikes are interspersed with each other which proves our hypothesis. 

However, when we decomposed OOVs into mixed-units (frequent word + single-letters), the WER dropped sharply to 10.17\% but still a little worse than the Vanilla CTC. This is again because of the mixture of words with single-letters. Next, we decomposed the OOVs into a combination of frequent words and double-letters. The WER dropped further to 9.58\%. When triple-letters and frequent words were used (totally 33k outputs), the WER dropped even more to 9.32\%. This is a 5.28\% relative WERR over Vanilla CTC. Then we applied attention on this model. To save computational costs, because of large number of output units, we excluded the PLM network in Eq.~\eqref{eq:CTCAttnLM-LSTM}. This model achieved a WER of 8.65\%, which is about 12.09\% relative WERR over the Vanilla CTC. This is our final word CTC model (mixed-units with triple-letters + attention). As an additional experiment, instead of mixed-units, we used wordpieces as targets. This model achieved a WER of 9.73\% which is a little better than that achieved with Vanilla CTC but worse than the results obtained with mixed-unit CTC. This indicates that building A2W models using mixed-units or WPMs is a better choice than simply using words and OOV (as in Vanilla CTC). 

Finally, we compared our final word CTC model with a traditional CD phoneme CTC in Table~\ref{Tab:WER_summary_wordCTC}. We trained a CD phoneme 6-layer BLSTM with the CTC criterion, modeling around 9000 tied CD phonemes. It has the same structure as other CTC models except that it uses different output units (phonemes instead of mixed-units or words). This CD phoneme CTC model achieved 9.28\% WER when decoding with a well-trained 5-gram LM with totally around 100 million (M) N-grams. Despite a strong CD phoneme CTC model and LM, the mixed-unit + Attention CTC model (without any LM or complex decoder) was still able to outperform it by about 6.79\% relative.

Note that the proposed model not only reduces the WER of the word CTC but also improves the end-user experience. The proposed model provides more meaningful outputs without outputting any OOV token which can be distracting to users. Moreover, we observed that even when the proposed model failed to recognize the OOVs accurately, it still came out with words which were a close match with the ground truth words. For example, the proposed method recognizes ``text fabine'' as ``text fabian'' and ``call zubiate'' as ``call zubiat''. However, the Vanilla CTC recognized these words as ``text OOV'' and ``call OOV'' respectively.

\section{Conclusions}
\label{sec: Conclusions}
We proposed improving letter and word CTC models using Attention CTC, Self-Attention CTC, Hybrid CTC, and Mixed-unit CTC. In attention-based CTCs, we generated new hidden features that carry attention weighted context information which are more useful than hidden features without context information. To solve the OOV issue in word CTC, we presented Hybrid CTC which uses a word and letter CTC as primary and auxiliary tasks in an MTL framework. Finally, to boost the performance of Hybrid CTC, we introduced Mixed-unit CTC whose output units contain both words and multi-letters. While the frequent words are treated as whole word units, the OOVs are decomposed into a sequence of frequent words and multi-letters. We evaluated all these methods on a 3400 hours Microsoft Cortana voice assistant task. The proposed word-based Mixed-unit CTC model with triple letters when combined with attention improved over the word-based Vanilla CTC model by 12.09\% relative. Such an acoustic-to-word CTC model is a pure end-to-end model without using any LM and complex decoder. It also outperformed a traditional CD phoneme CTC model equipped with strong LM and complex decoder by 6.79\% relative.

\section{Code}
\label{sec: Code}
The CNTK script for Attention CTC described in Section~\ref{sec: CTCAttn} is available online at: \url{https://github.com/microsoft/CNTK/tree/vadimma/CTC/Examples/Speech/AttentionCTC}.

\section*{Acknowledgment}
The authors would like to thank Kastubh Kalgaonkar, during his time at Microsoft, for his help in building WPMs.

\ifCLASSOPTIONcaptionsoff
  \newpage
\fi



\bibliographystyle{IEEEtran}
\bibliography{strings,refs}

\end{document}